\documentclass[sigconf]{acmart}
\AtBeginDocument{%
  }

\copyrightyear{2026}
\acmYear{2026}
\setcopyright{cc}
\setcctype{by}
\acmConference[MM '26]{Proceedings of the 34th ACM International
  Conference on Multimedia}{November 10--14, 2026}{Rio de Janeiro, Brazil}
\acmBooktitle{Proceedings of the 34th ACM International Conference on
  Multimedia (MM '26), November 10--14, 2026, Rio de Janeiro, Brazil}
\acmDOI{10.1145/3767308.3837701}
\acmISBN{979-8-4007-2213-4/2026/11}

\usepackage{tcolorbox}
\usepackage{listings}
\usepackage{pifont}
\usepackage{subcaption}
\usepackage{float}
\usepackage{marvosym}

\newcommand{\cmark}{\ding{51}}
\newcommand{\xmark}{\ding{55}}
\definecolor{codegray}{rgb}{0.95,0.95,0.95}
\lstdefinestyle{reportcode}{
  basicstyle=\ttfamily\fontsize{6.5}{7.5}\selectfont,
  breaklines=true,
  columns=fullflexible,
  frame=none,
  xleftmargin=0pt,
  xrightmargin=0pt,
  aboveskip=0pt,
  belowskip=0pt
}
\begin{document}

\title{SEED: \underline{S}imple ViT and \underline{E}volving Harness \\ for \underline{E}xplainable Text Forgery \underline{D}etection}

\author{Kahim Wong}
\email{yc37437@um.edu.mo}
\affiliation{%
  \institution{University of Macau}
  \city{Macau}
  \country{China}
}

\author{Kemou Li}
\email{kemou.li@connect.umac.mo}
\affiliation{%
  \institution{University of Macau}
  \city{Macau}
  \country{China}
}

\author{Yiming Chen}
\email{yc17486@um.edu.mo}
\affiliation{%
  \institution{University of Macau}
  \city{Macau}
  \country{China}
}

\author{Haiwei Wu}
\email{haiweiwu@uestc.edu.cn}
\affiliation{%
  \institution{University of Electronic Science and Technology of China}
  \city{Chengdu}
  \country{China}
}

\author{Jiantao Zhou}
\authornote{Corresponding author}
\email{jtzhou@um.edu.mo}
\affiliation{%
  \institution{University of Macau}
  \city{Macau}
  \country{China}
}

\renewcommand{\shortauthors}{Wong et al.}

\begin{abstract}
AI-assisted image editing threatens trust in financial, legal, and identity records. The GenText-Forensics Challenge at ACM MM 2026 addresses this by requiring structured forensic reports, in which integrating detection, pixel-level localization, and natural language explanation for multilingual text-centric forgery images. We present \textbf{SEED}, a modular system with three components. First, a similarity-guided pipeline augments training with diverse synthetic forgeries. Second, a single ViT, built on DINOv3 with LoRA adaptation, jointly performs detection and pixel-level localization while preserving pre-trained priors with minimal trainable parameters. Third, an evolving harness takes the detector's predictions and generates a complete forensic report via an MLLM, iteratively improved through a proposer-evaluator loop optimizing report quality. SEED ranked 3rd in the GenText-Forensics Challenge. Code and data are available at \url{https://github.com/KahimWong/GenText-Forensics-3rd-Place}.
\end{abstract}

\begin{CCSXML}
<ccs2012>
   <concept>
       <concept_id>10010147.10010178.10010224.10010245.10010247</concept_id>
       <concept_desc>Computing methodologies~Image segmentation</concept_desc>
       <concept_significance>300</concept_significance>
       </concept>
 </ccs2012>
\end{CCSXML}

\ccsdesc[300]{Computing methodologies~Image segmentation}

\keywords{Image Forgery Localization, Document Forensics, Vision Transformer, LoRA, Large Language Model, Meta-Harness}

\maketitle

\section{Introduction}

The widespread availability of AI-powered image editing tools has fundamentally lowered the barrier to image manipulation, enabling realistic text-level forgeries at scale~\cite{wong2025fontguard,wong2025end,wong2026k,rombach2022high,suvorov2022resolution,tuo2024anytext,ju2024brushnet, li2026llm, wu2026editprint, wu2026zero}. Detecting such forgeries is challenging, since the tampered text always seamlessly blends into structured layouts and forged regions lack the telltale boundary artifacts common in natural images~\cite{qu2023towards,wong2025adcd}. These challenges demand detectors that not only localize manipulation but also produce \emph{explainable} evidence, a capability largely absent in prior work. \\

The GenText-Forensics Challenge at ACM Multimedia 2026~\cite{gentext2026} addresses this emerging threat through a novel formulation. Beyond image-level detection and pixel-level localization, systems must generate structured forensic reports that explain \emph{what} was manipulated, \emph{where} the manipulation occurred, and \emph{why} the evidence supports a forgery conclusion. We describe the task, dataset, and evaluation protocol in Section~\ref{sec:task}. \\

In this technical report, we present \textbf{SEED}, our 3rd-place solution in the GenText-Forensics Challenge, which decomposes the forgery analysis task into three complementary modules. First, we augment the training data through a similarity-guided synthetic forgery generation pipeline (Sec.~\ref{sec:synth}) that produces realistic document forgeries across five manipulation types. We further introduce a clean-forged paired training strategy that encourages discriminative learning by contrasting authentic and manipulated versions of the same image. Second, we design a ViT-based forgery detector (Sec.~\ref{sec:detector}) that adapts the DINOv3 ViT-L/16 \cite{simeoni2025dinov3} backbone with LoRA adaptation \cite{hu2022lora}. We leverage the EoMT \cite{kerssies2025your} structure \cite{cheng2022masked} that turns the ViT into a localization model with minimal additional parameters. By freezing most pre-trained parameters and adapting only low-rank updates, SEED's detector preserves transferable visual priors while learning forgery-specific traces with minimal additional parameters. Third, we employ the Meta-Harness~\cite{lee2026meta} (Sec.~\ref{sec:harness}) that iteratively evolves MLLM harnesses through a proposer-evaluator loop, yielding progressively better forensic reports. Our main contributions are as follows.
\begin{itemize}
    \item A simple yet effective forgery model based on DINOv3 ViT backbone with LoRA adaptation and an EoMT structure that unifies image-level detection and pixel-level localization with minimal additional parameters.
    \item A training strategy, paired clean-forgery batch construction, that significantly boosts detection performance by forcing the model to contrast authentic and manipulated versions of the same source image.
    \item A Meta-Harness approach that automatically discovers effective harness for forensic report generation without manual prompt engineering.
    \item A pipeline integrating contrastive-guided synthetic data generation, ViT detector, and evolving harness, achieving 3rd place in the GenText-Forensics Challenge.
\end{itemize}

\section{Related Work}

\subsection{Text-Centric Image Forgery Localization}

Text-centric image forgery localization (TFL) has progressed from early benchmarking efforts to increasingly robust detectors. DTD \cite{qu2023towards} introduced the DocTamper benchmark and demonstrated that JPEG DCT coefficient analysis combined with multi-scale decoding can effectively localize tampered regions in document images. FFDN \cite{chen2024enhancing} improved localization through RGB-DCT fusion with feature enhancement modules. ADCD-Net \cite{wong2025adcd} explicitly addressed the misalignment between DCT grid boundaries and forgery regions, and introduced adaptive text-background disparity modeling. TIFDM \cite{dong2024robust} strengthened trace enhancement and multi-scale aggregation under JPEG compression degradations. CAFTB \cite{song2025cross} combined spatial and noise-domain cues through cross-attention fusion. However, most existing TFL detectors rely on full-parameter fine-tuning (FPFT) of custom CNN-Transformer architectures. Recent work \cite{yan2024orthogonal} has shown that FPFT can induce low-rank feature collapse in vision foundation models, harming cross-domain generalization. LoRA \cite{hu2022lora} preserves pre-trained priors by freezing dominant weight components and learning only low-rank updates, achieving strong performance in NLP and vision tasks with minimal trainable parameters. SEED's detector builds on this insight by applying LoRA to a DINOv3 ViT backbone for TFL.

\subsection{Synthetic Data for Document Forensics}

Data augmentation through synthetic forgery generation has become essential for training robust forensic detectors. Earlier work such as DTD~\cite{qu2023towards} already demonstrated the value of large-scale synthetic tampering for document forgery localization. Recent work~\cite{dhouib2026leveraging} further improves synthetic forgery generation using contrastively trained models to guide crop selection. A crop-similarity model measures semantic compatibility between candidate source and target regions, while a crop-quality model evaluates the visual fidelity of inserted crops. This approach produces realistic forgeries across five manipulation types with natural text-background blending. We adopt this method to generate paired real-synthetic training data for the GenText-Forensics challenge.

\subsection{LLM-Based Forensic Report Generation}

The use of large language models for structured forensic reporting is an emerging area. Recent work such as TextShield-R1~\cite{qu2026textshield} shows that MLLMs can be trained to perform tampered text detection, localization, and reasoning in an end-to-end manner. The Meta-Harness framework \cite{lee2026meta} instead introduces an automated search over prompt strategies, visual representations, and output repair logic using a proposer-evaluator loop, eliminating manual prompt engineering. We adapt this framework to the forgery analysis domain, where the harness must reason over predicted forgery masks, construct visual overlays, and produce reports matching a strict forensic schema.

\section{Task and Dataset}
\label{sec:task}

\begin{figure}[h]
\centering
\begin{subfigure}[t]{0.48\linewidth}
\centering
\includegraphics[width=\linewidth]{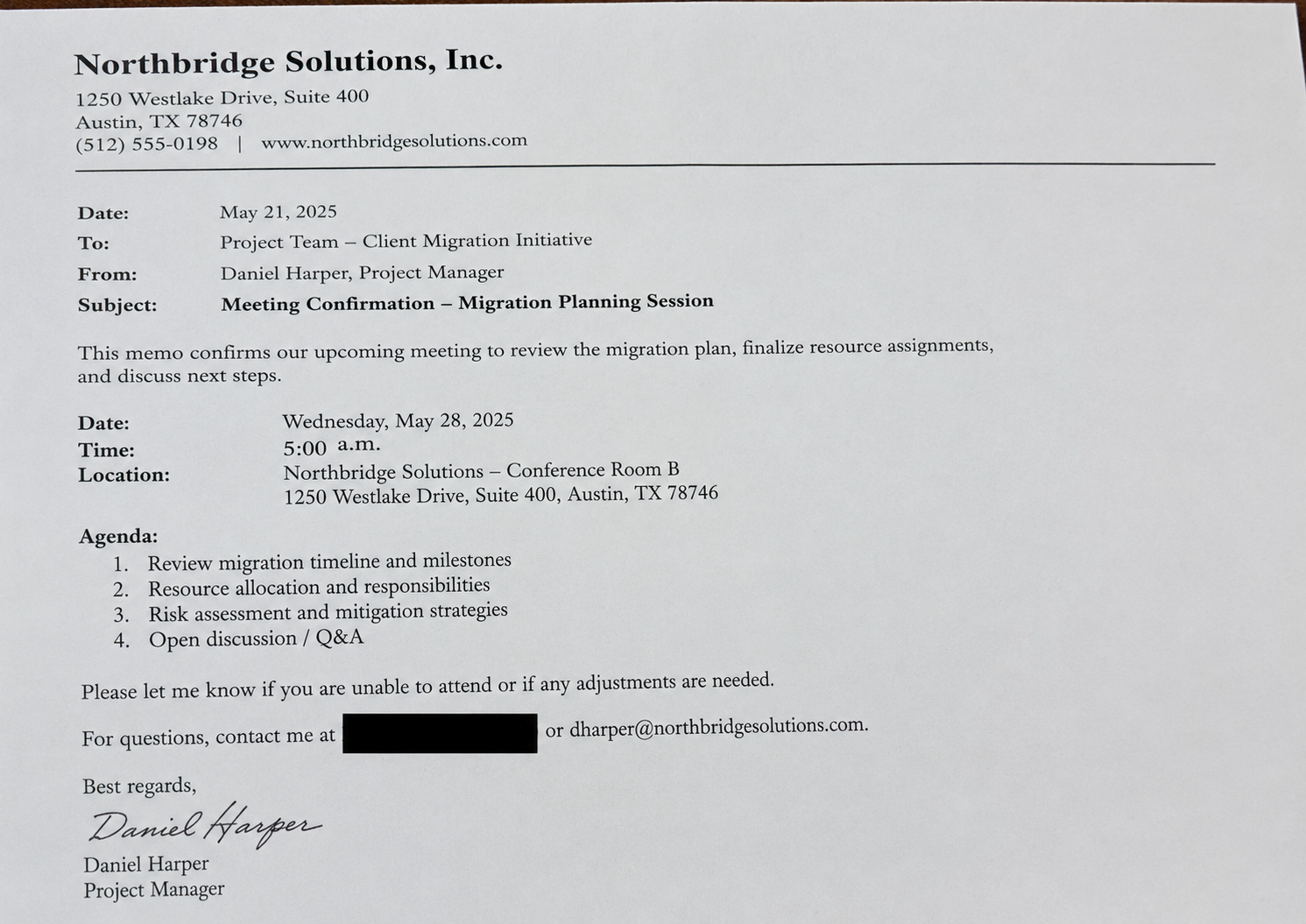}
\caption{Forged Image}
\end{subfigure}
\hfill
\begin{subfigure}[t]{0.48\linewidth}
\centering
\includegraphics[width=\linewidth]{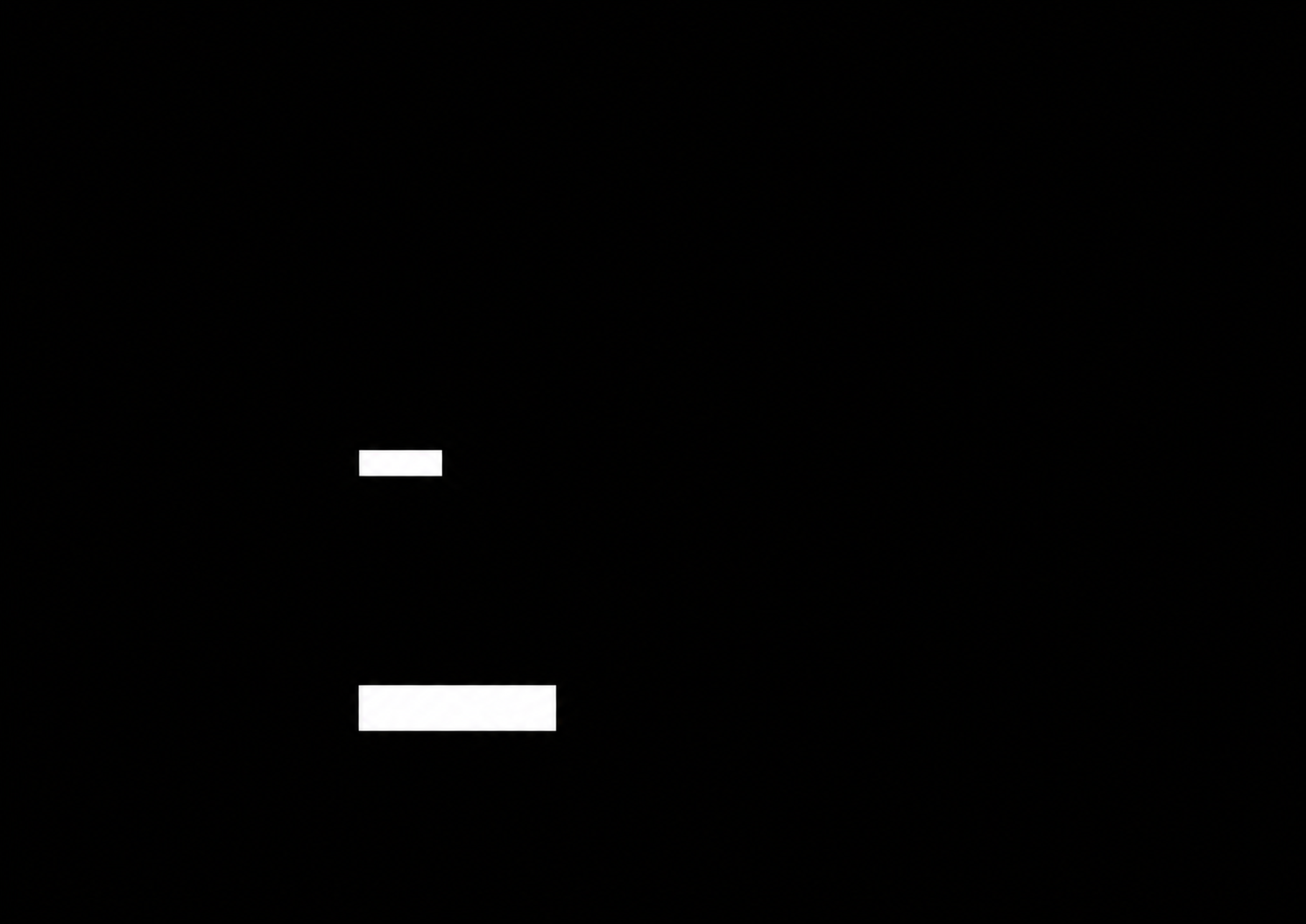}
\caption{Forged Mask}
\end{subfigure}

\vspace{6pt}

\begin{subfigure}[t]{\linewidth}
\centering
\begin{tcolorbox}[
    colback=codegray,
    colframe=black!75,
    arc=2pt,
    boxrule=0.5pt,
    left=4pt,
    right=4pt,
    top=4pt,
    bottom=4pt
]
\begin{lstlisting}[style=reportcode]
# FORGERY ANALYSIS REPORT
**Overall Assessment:**
    **[Conclusion]:** FORGED
    **[RISK_SCORE]:** 73
---
## DETAILED ANOMALY ANALYSIS
### ANOMALY_001: Visual Clumsy Alteration
[GROUNDING]: [1081, 933, 1288, 998]
[REASON]: A crude black block obscures the phone number; sharp edges and uniform color create a clear discontinuity.
### ANOMALY_002: Logical Fraud
[GROUNDING]: [1372, 585, 1630, 655]
[REASON]: The meeting time '5:00 a.m.' contradicts standard business practice, and the 'a.m.' glyph shows misalignment.
---
## SUMMARY
The document exhibits 2 distinct anomalies involving crude redaction and logical inconsistency.
**END OF REPORT**
\end{lstlisting}
\end{tcolorbox}
\caption{Forensic Report}
\end{subfigure}
\caption{Example with (a) the forged image, (b) its forgery mask, and (c) the target forensic report in the required Markdown format.}
\label{fig:report_example}
\end{figure}

The GenText-Forensics Challenge~\cite{gentext2026} formulates document forgery analysis as a unified generative task. Given a text-centric image, e.g. Fig.~\ref{fig:report_example}~(a), the system produce a structured forensic report that integrates three capabilities (detection, localization, and explanation). Detection answers whether the document is forged, localization identifies where the manipulated regions are, and explanation provides the evidence that supports the conclusion. The target report format follows a strict Markdown schema, as illustrated in Fig.~\ref{fig:report_example}~(c).

\begin{figure*}[t]
    \centering
    \includegraphics[width=\linewidth]{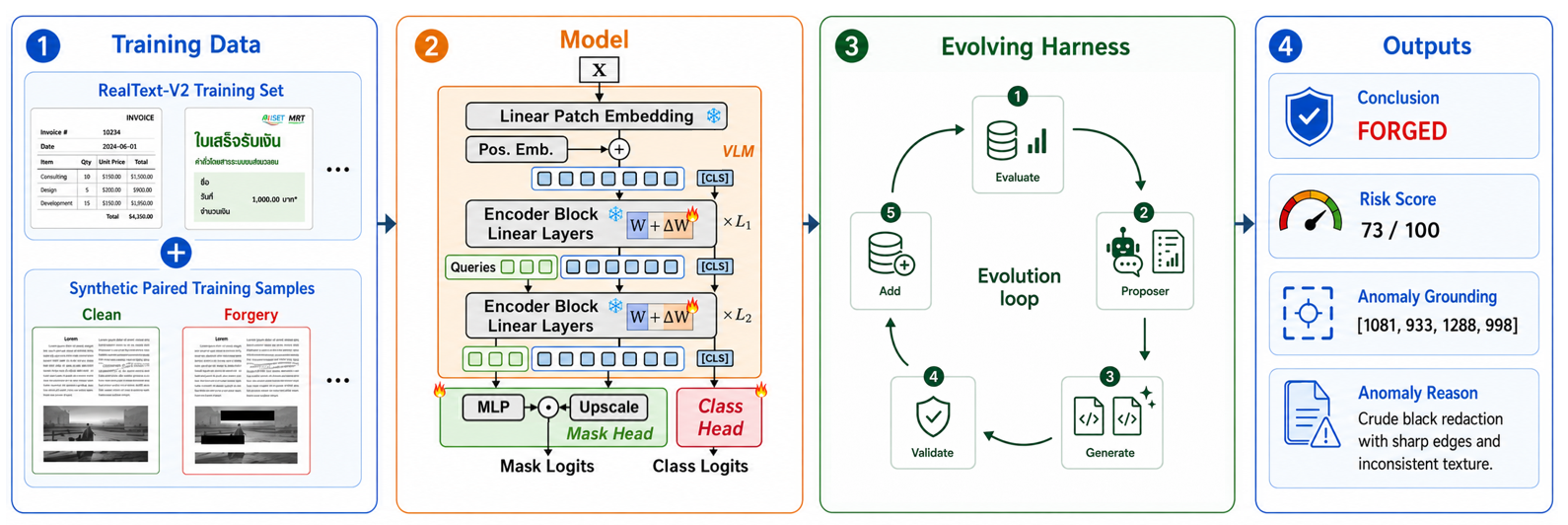}
    \caption{Overview of SEED's three-stage forgery analysis pipeline. Stage~1 generates diverse synthetic forgeries using contrastive-guided crop selection. Stage~2 detects and localizes forged regions with a DINOv3 ViT. Stage~3 an evolving loop automatically discovers effective harness for converting raw ViT outputs into structured forensic reports. Stage~4 produces structured forensic reports through the final evolved MLLM harness.}
    \label{fig:overview}
\end{figure*}

The evaluation metric combines detection, localization, and explanation quality into a single final score. Detection quality is measured by image-level F1 and denoted $S_{\text{Det}}$. Localization quality is measured by mask mIoU and denoted $S_{\text{Loc}}$. Explanation quality is measured by BERTScore and denoted $S_{\text{Exp}}$. Report quality is measured by an LLM judge rubric and denoted $S_{\text{Rep}}$. The final score:

\begin{equation}
\label{eq:sfin}
S_{\text{Fin}} = 0.3\,S_{\text{Det}} + 0.4\,S_{\text{Loc}} + 0.15\,S_{\text{Exp}} + 0.15\,S_{\text{Rep}}.
\end{equation}

The challenge is built on RealText-V2, a large-scale multilingual document forgery benchmark with 20K+ samples across 6 languages and 6 domains (finance, healthcare, education, legal, identity, general). Each sample includes a document image, a pixel-level binary forgery mask, a forgery-type label, and an expert-authored forensic report. The training set covers 100+ manipulation methods, from character-level substitution to sentence-level semantic edits. Participants are prohibited from using external datasets and must submit a single file of structured Markdown reports for all test images.

\section{Method}
\label{sec:method}

Our solution \textbf{SEED} consists of three stages: (1) synthetic forgery data generation, (2) forgery detection and localization using a ViT with LoRA adaptation, and (3) MLLM-based forensic report generation through a Meta-Harness loop (Figure~\ref{fig:overview}).

\subsection{Synthetic Forgery Data Generation}
\label{sec:synth}

To increase forgery diversity beyond the original 
RealText-V2 training set, we adopt a similarity-guided synthetic method~\cite{dhouib2026leveraging} that generates forgeries from the clean images within the RealText-V2 samples. Specifically, the synthetic method uses two trained selection and quality models to automatically select source-target crop pairs from these clean images and produce high-quality forgeries across five manipulation types, including copy-move, splicing, insertion, inpainting, and coverage. We combine these synthetic pairs with the original RealText-V2 samples for joint training. Table~\ref{tab:data_composition} summarizes the size of the original RealText-V2 training set and our synthetic set. Each synthetic forged sample is paired with its pristine source image, forming a (clean, forged) training pair. We construct each training batch by sampling matched clean-forgery pairs, so the model jointly processes both the original document and its manipulated counterpart.

\begin{table}[t]
    \caption{Training data composition for detector learning.}
    \label{tab:data_composition}
    \centering
    \normalsize
    \setlength{\tabcolsep}{5pt}
    \begin{tabular}{llr}
        \toprule
        Source & Type & \# Samples \\
        \midrule
        RealText-V2 & authentic & 6000 \\
        RealText-V2 & forged & 7500 \\
        RealText-V2 & Total & 13500 \\
        \midrule
        Synthetic & copy-move & 5009 \\
        Synthetic & coverage & 5002 \\
        Synthetic & inpainting & 5004 \\
        Synthetic & insertion & 5004 \\
        Synthetic & splicing & 5009 \\
        Synthetic & Total & 25028 \\
        \bottomrule
    \end{tabular}
\end{table}

\subsection{Forgery Detection Model}
\label{sec:detector}

SEED's forgery detector is designed around three principles. (1) Preserve transferable visual priors from visual foundation model (e.g. DINOv3), (2) learn forgery-specific traces through minimal parameter adaptation, and (3) efficiently handle both image-level detection and pixel-level localization in a unified architecture.

\paragraph{Architecture}

We adopt DINOv3 ViT-L/16 \cite{simeoni2025dinov3} as a frozen backbone to preserve the transferable visual priors. To capture forgery-specific traces with minimal parameter overhead, we apply LoRA \cite{hu2022lora} with $r=1$ to the query, key, value, and output projections of the self-attention layers, as well as to the up and down projections of the MLP layers in all the ViT blocks, while keeping all other backbone parameters frozen. For unified image-level detection and pixel-level localization, we prepend a \texttt{[CLS]} token and feed its final representation into a classification head for forgery detection, and adapt the EoMT framework \cite{kerssies2025your} with a Mask2Former-style mask head for pixel-level localization.

\begin{figure}[tb]
    \centering
    \includegraphics[width=0.78\linewidth]{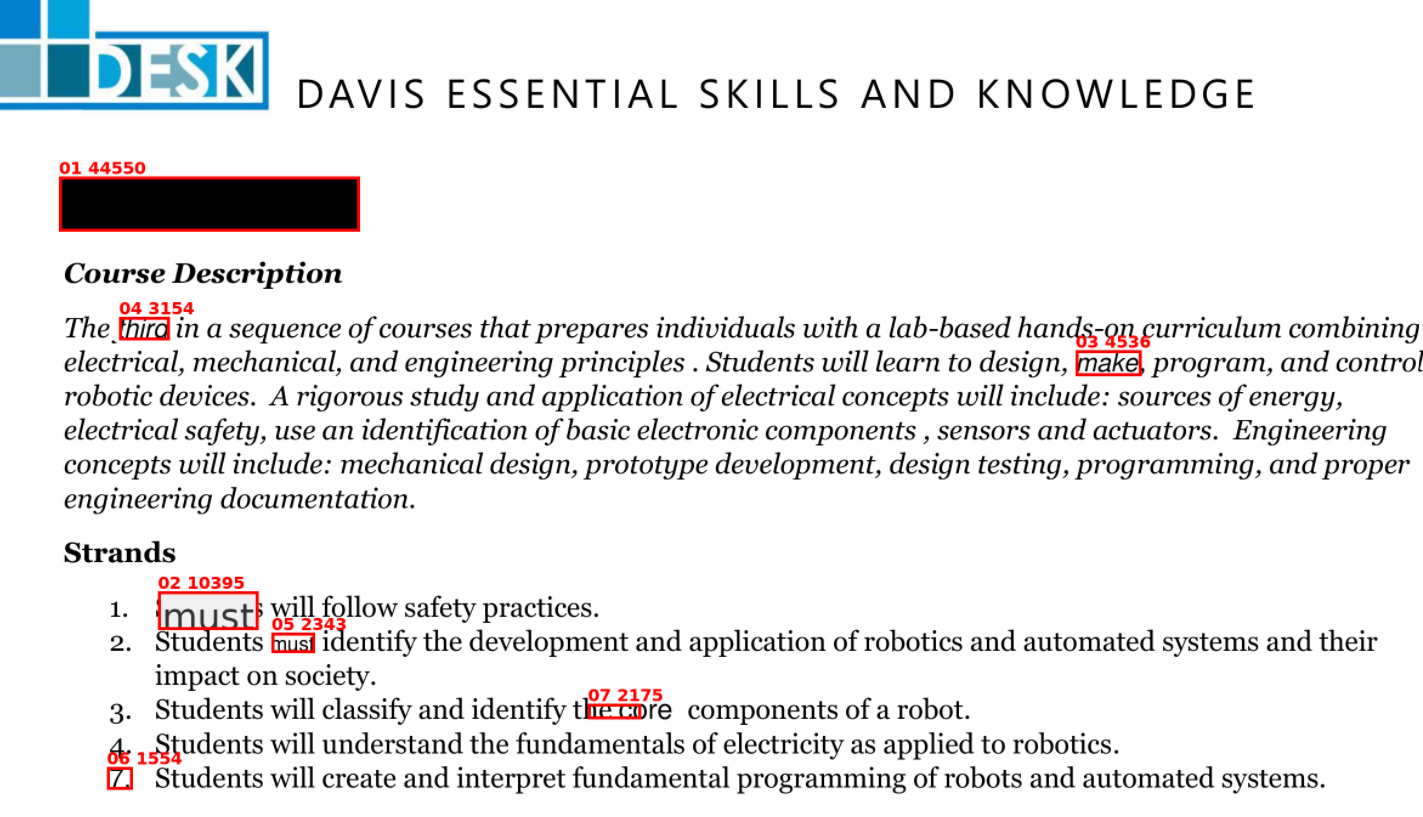}
    \caption{Visual input to the MLLM in the harness stage: the predicted forgery mask is overlaid on the document so the MLLM can reason about suspicious regions before generating the report.}
    \label{fig:overlap_input}
\end{figure}  

Specifically, the model receives an input image $\mathbf{X} \in \mathbb{R}^{3 \times H \times W}$ and produces two outputs, an image-level forgery probability $\hat{y} \in [0,1]$ and a pixel-level forgery map $\hat{\mathbf{P}} \in [0,1]^{H \times W}$. The ViT backbone first encodes $\mathbf{X}$ into patch tokens, which are processed through $L$ transformer blocks. We prepend an additional \texttt{[CLS]} token to the patch sequence prior to the first block to aggregate global forgery cues. After all $L$ blocks, the \texttt{[CLS]} token's final representation is passed through a classification head $\phi_{\text{cls}}$, implemented as a linear projection followed by softmax, producing the forgery probability $\hat{y}$. For localization, following EoMT we insert a learnable query token before the final $L_2 = 4$ ViT blocks. After these blocks, the query token yields query embeddings $\mathbf{Q} \in \mathbb{R}^{N_q \times d}$ and the patch tokens yield dense features $\mathbf{Z} \in \mathbb{R}^{H_p \times W_p \times d}$. The mask head follows Mask2Former \cite{cheng2022masked}. The query features are transformed via an MLP $\phi_{\text{mlp}}$, the dense features are upsampled via $\phi_{\text{up}}$, and the two are combined via dot product to produce mask logits $\mathbf{M} \in \mathbb{R}^{N_q \times H_p \times W_p}$. These logits are bilinearly upsampled to the input resolution and passed through a sigmoid to obtain the final probability map $\hat{\mathbf{P}}$.

\paragraph{Paired clean-forgery training.} A key design choice is how training batches are constructed for the synthetic samples. Since each synthetic forgery is generated from a known clean document image, we can explicitly pair them within the same batch. Rather than randomly shuffling authentic and forged images, we construct each batch with matched (clean, forged) pairs of the same document. This forces the model to observe both versions of the same content under identical optimization steps. As shown in our experiments, this simple paired strategy consistently outperforms standard random-shuffle training.

\paragraph{Training Objective}

The model is trained end-to-end with a composite loss combining Mask2Former losses. For each training sample $(\mathbf{X}, \mathbf{Y}, y)$ where $\mathbf{Y} \in \{0,1\}^{H \times W}$ is the ground-truth mask and $y \in \{0,1\}$ indicates authenticity:
\begin{equation}
\mathcal{L} = \lambda_{\text{CE}}\mathcal{L}_{\text{CE}}(\mathbf{c}, y) + \lambda_{\text{BCE}}\mathcal{L}_{\text{BCE}}(\hat{\mathbf{P}}, \mathbf{Y}) + \lambda_{\text{Dice}}\mathcal{L}_{\text{Dice}}(\hat{\mathbf{P}}, \mathbf{Y})
\end{equation}
where $\lambda_{\text{CE}} = 1.0$, $\lambda_{\text{BCE}} = 5.0$, $\lambda_{\text{Dice}} = 5.0$.

\subsection{Evolving Harness for Report Generation}
\label{sec:harness}

As shown in Figure~\ref{fig:overview} (Stage~3), the final step transforms the detector's raw outputs into a structured forensic report matching the schema in Figure~\ref{fig:report_example}. This conversion is non-trivial because the detector produces an image-level forgery probability $\hat{y}$ and a pixel-level probability map $\hat{\mathbf{P}}$, while the challenge requires a natural-language report with binary verdicts, localized bounding-box groundings, forensic reasoning for each anomaly, and a summary. The two representations are fundamentally different. Rather than manually engineering prompts and repair logic for this cross-modal bridge, we employ a Meta-Harness framework~\cite{lee2026meta} that \emph{automatically evolves} harness candidates through a proposer-evaluator loop. Each harness candidate is a self-contained module encapsulating mask rendering, mask to bounding boxes, prompt construction and output repair scripts. Figure~\ref{fig:overlap_input} shows the visual input passed to the MLLM, where the predicted forgery mask is overlaid on the document image to make suspicious regions explicit before report generation.

\paragraph{Initialization.} Following the Meta-Harness onboarding protocol\footnote{https://github.com/stanford-iris-lab/meta-harness}, a coding agent (e.g.\ Opencode) was pointed to \texttt{ONBOARDING.md} and, through a structured conversation, produced a \texttt{domain\_spec.md} defining the harness interface, metrics, search budget, and baseline strategy. The same agent then implemented the full framework from scratch: the search loop, seed harnesses, a proposer, and an evaluator. No manual code or prompt engineering was performed.

\begin{table*}[t]
    \caption{Cross-domain localization and detection F1 scores. All models use a DINOv3 ViT-L/16 backbone with LoRA adaptation unless noted. \#: the configuration index; LoRA: the LoRA rank; Step: training steps; Batch: batch size; JPEG: whether JPEG augmentation is applied; Data: the training data split; and Paired: whether matched clean-forgery batches are used.}
    \label{tab:loc_results}
    \centering
    \normalsize
    \setlength{\tabcolsep}{2pt}
    \begin{tabular}{c|lccccc|ccccc|ccccc}
        \toprule
        \multicolumn{7}{c|}{Configuration} & \multicolumn{5}{c|}{Localization F1} & \multicolumn{5}{c}{Detection F1} \\
        \# & LoRA & Step & Batch & JPEG & Data & Paired & T-SROIE & OSTF & TPIC-13 & RTM & Avg & T-SROIE & OSTF & TPIC-13 & RTM & Avg \\
        \midrule
        1 & r=32 & 10k & 60 & \cmark & Train & \xmark & 0.738 & 0.441 & 0.636 & 0.126 & 0.485 & 0.681 & 0.259 & 0.424 & 0.162 & 0.381 \\
        2 & r=32 & 10k & 20 & \cmark & Train & \xmark & 0.734 & 0.533 & 0.699 & 0.150 & 0.529 & 0.713 & 0.450 & 0.685 & 0.160 & 0.502 \\
        3 & r=32 & 5k & 20 & \cmark & Train & \xmark & 0.661 & 0.613 & 0.736 & 0.145 & 0.539 & 0.745 & 0.607 & 0.750 & 0.195 & 0.574 \\
        4 & r=1 & 5k & 20 & \cmark & Train & \xmark & 0.764 & 0.605 & 0.721 & 0.165 & 0.564 & \textbf{0.784} & 0.648 & 0.751 & 0.159 & 0.585 \\
        5 & r=1 (MLP) & 5k & 20 & \cmark & Train & \xmark & 0.756 & 0.627 & 0.730 & 0.161 & 0.568 & 0.648 & 0.602 & 0.729 & 0.161 & 0.535 \\
        6 & r=1 & 5k & 20 & \xmark & Train & \xmark & 0.745 & 0.640 & 0.747 & 0.156 & 0.572 & 0.720 & 0.665 & 0.799 & 0.196 & 0.595 \\
        7 & r=1 & 2k & 20 & \xmark & Train & \xmark & 0.754 & 0.615 & 0.697 & 0.177 & 0.561 & 0.459 & 0.681 & 0.789 & 0.178 & 0.527 \\
        8 & r=1 & 5k & 20 & \xmark & Train+Syn & \xmark & \textbf{0.800} & 0.711 & \textbf{0.810} & \textbf{0.178} & \textbf{0.625} & 0.659 & 0.810 & 0.916 & \textbf{0.210} & 0.649 \\
        9 & r=1 & 5k & 20 & \xmark & Train+Syn & \cmark & 0.782 & \textbf{0.718} & 0.798 & \textbf{0.178} & 0.619 & 0.738 & \textbf{0.832} & \textbf{0.930} & 0.207 & \textbf{0.677} \\
        \bottomrule
    \end{tabular}%
\end{table*}

\paragraph{Harness interface.} Each harness implements a uniform interface. It receives the original image, the predicted mask and image-level forgery probability, renders a visual overlay of predicted forgeries on the image, constructs a prompt for the base MLLM, calls the MLLM, repairs the output to enforce schema compliance with mandatory tags
\texttt{[Conclusion]}, \texttt{[RISK\_SCORE]}, \texttt{[GROUNDING]},
\texttt{[REASON]}, and \texttt{END OF REPORT}, and returns a valid
forensic report.

\paragraph{Evaluation.} Each candidate is evaluated on a fixed search set of 50 training samples. Two scores are computed separately. First, $S_{\text{Exp}}$ is the BERTScore F1 between the generated explanation text (all \texttt{[REASON]} sections plus the \texttt{SUMMARY}) and the ground-truth expert report. Second, $S_{\text{Rep}}$ is an LLM-judge rubric score (0--100) produced by GPT-4o-mini evaluating factuality, reasoning quality, and completeness of the generated report. The composite score is $S = 0.15\,S_{\text{Exp}} + 0.15\,S_{\text{Rep}}$, which corresponds to the explanation-related terms in Eq.~\eqref{eq:sfin}.

\paragraph{Evolution loop.} Each iteration proceeds through five steps. (1) Evaluate all current candidates on the search set. (2) Construct a proposer prompt containing the Pareto-frontier scores, representative failure cases, and the source code of the best-performing harnesses. (3) The proposer LLM generates 2 new harness candidates, each with a stated hypothesis about what improvement it introduces. (4) New candidates are validated for import correctness and interface compliance. (5) Valid candidates are evaluated and added to the population. The loop runs for $T = 30$ iterations, maintaining a Pareto frontier of non-dominated candidates across $S_{\text{Exp}}$ and $S_{\text{Rep}}$.

\paragraph{Final selection.} After the search budget is exhausted, the top-performing harness on the search set is selected to produce the final submission reports. The entire harness code is auto-generated by the proposer LLM. No manual tuning of prompts, few-shot examples, or repair logic is performed. This ensures that the explanation component is itself the product of a principled, reproducible optimization process.

\section{Experiments}
\label{sec:experiments}

\subsection{Dataset and Evaluation}

We use the RealText-V2 training set as described in Section~\ref{sec:task}. For ablation studies, we additionally report pixel-level localization F1, precision, and recall, as well as image-level detection F1. We evaluate on the original RealText-V2 training set and the synthetic forgery training set with random 1000 samples for in-domain evaluation. We also evaluate on four cross-domain test sets from ForensicHub~\cite{du2025forensichub}: T-SROIE~\cite{wang2022tampered} (scanned receipts with AIGC text editing), OSTF~\cite{qu2025revisiting} (natural scene text with 8 AIGC models), TPIC-13~\cite{wang2022detecting} (scene-text images with SR-Net editing), and RTM~\cite{luo2025toward} (mixed synthetic and manual document manipulations). All ForensicHub samples are cropped to $512 \times 512$ resolution.

\subsection{Implementation Details}

\begin{table*}[t]
    \caption{Effect of training image size under patched detection-first inference. Train-1000 and Syn-1000 denote random 1000-sample subsets drawn from the training set and synthetic set, respectively. Each row reports the best threshold found for that training image size.}
    \label{tab:imgsize_results}
    \centering
    \normalsize
    \setlength{\tabcolsep}{3pt}
    \begin{tabular}{ccc|cccccc|cccccc}
        \toprule
        & & & \multicolumn{6}{c|}{Train-1000} & \multicolumn{6}{c}{Syn-1000} \\
        Img size & Det-thr & Mask-thr & Pix F1 & Pix P & Pix R & Img F1 & Img P & Img R & Pix F1 & Pix P & Pix R & Img F1 & Img P & Img R \\
        \midrule
        512 & 0.99 & 0.99 & 0.673 & 0.889 & 0.541 & 0.992 & 0.986 & 0.998 & 0.636 & \textbf{0.982} & 0.470 & 0.912 & 0.984 & 0.850 \\
        786 & 0.99 & 0.99 & 0.700 & 0.909 & 0.570 & 0.991 & 0.986 & 0.996 & 0.673 & 0.976 & 0.514 & 0.924 & 0.984 & 0.870 \\
        1024 & 0.99 & 0.99 & 0.720 & 0.906 & 0.598 & 0.995 & 0.996 & 0.994 & 0.682 & 0.971 & 0.526 & 0.917 & 0.995 & 0.850 \\
        1280 & 0.99 & 0.99 & 0.737 & \textbf{0.941} & 0.605 & 0.994 & \textbf{0.998} & 0.990 & 0.694 & 0.968 & 0.541 & 0.924 & \textbf{0.998} & 0.860 \\
        1280 & 0.95 & 0.95 & \textbf{0.761} & 0.892 & \textbf{0.664} & \textbf{0.996} & 0.994 & 0.998 & \textbf{0.725} & 0.946 & \textbf{0.588} & \textbf{0.931} & 0.993 & \textbf{0.876} \\
        \bottomrule
    \end{tabular}%
\end{table*}

\paragraph{Detector training.} We train the detector for 5k steps using AdamW optimizer with learning rate $3 \times 10^{-4}$ decayed to $1 \times 10^{-5}$ via cosine annealing, weight decay $1 \times 10^{-4}$, and batch size 20. Training uses automatic mixed precision (FP16), and is distributed across 5 NVIDIA RTX 3090 GPUs using DDP.

\paragraph{Harness configuration.} The base MLLM is Qwen3.5-Flash, the judge model is GPT-4o-mini, and the proposer is GPT-5.5. The Meta-Harness search runs for 30 iterations on a fixed subset of 50 training samples, with 2 candidates per iteration. BERTScore uses the \texttt{google-bert/bert-base-multilingual-cased} model.

\subsection{Detection and Localization Results}

\begin{table}[tb]
    \caption{Threshold ablation for detection-first inference on 1000 training samples. Results use the Table~\ref{tab:loc_results} \#9 setting.}
    \label{tab:det_results}
    \centering
    \normalsize
    \setlength{\tabcolsep}{4pt}
    \begin{tabular}{cc|cccccc}
        \toprule
        Det-thr & Mask-thr & Pix F1 & Pix P & Pix R & Img F1 & Img P & Img R \\
        \midrule
        0.50 & 0.50 & 0.444 & 0.314 & \textbf{0.760} & 0.853 & 0.743 & \textbf{1.000} \\
        0.75 & 0.50 & 0.451 & 0.323 & 0.752 & 0.930 & 0.870 & \textbf{1.000} \\
        0.90 & 0.90 & 0.623 & 0.571 & 0.685 & 0.962 & 0.926 & \textbf{1.000} \\
        0.95 & 0.95 & 0.667 & 0.690 & 0.646 & 0.974 & 0.949 & \textbf{1.000} \\
        0.99 & 0.99 & \textbf{0.673} & 0.889 & 0.541 & \textbf{0.992} & 0.986 & 0.998 \\
        0.999 & 0.999 & 0.487 & \textbf{0.980} & 0.324 & 0.979 & \textbf{1.000} & 0.958 \\
        \bottomrule
    \end{tabular}%
\end{table}

Table~\ref{tab:loc_results} reports cross-domain localization and detection results. We analyze the effects of the training settings below.
    
\paragraph{Training steps, batch size, and LoRA rank.} Within the $r{=}32$ configurations, reducing capacity along all three axes improves cross-domain generalization, and the best results come from the lower-capacity $r{=}1$ settings. Lowering the LoRA rank from $r{=}32$ to $r{=}1$ (row~\#3 vs.\ \#6) lifts Avg Loc-F1 by $+6\%$ and Avg Det-F1 by $+4\%$; the best $r{=}1$ settings (\#8, \#9) reach $+16\%$ and $+18\%$ over the best $r{=}32$ baseline (\#3). Halving the steps from 10k to 5k at $r{=}32$ (row~\#2 vs.\ \#3) gains $+2\%$ Loc-F1 and $+14\%$ Det-F1, and reducing the batch from 60 to 20 (row~\#1 vs.\ \#2) gains $+32\%$ Det-F1 and $+9\%$ Loc-F1. Excess capacity thus overfits training-set patterns, consistent with low-rank adaptation better preserving pre-trained priors \cite{yan2024orthogonal}.

\paragraph{Attention + MLP vs. MLP-only LoRA.} Adapting only the MLP projections (row~\#5, $r{=}1$ MLP-only) versus adapting both attention and MLP projections (row~\#6, $r{=}1$ full) drops Avg Det-F1 from 0.595 to 0.535 ($-10\%$) while Avg Loc-F1 is nearly unchanged (0.572 vs.\ 0.568). Attention-layer adaptation is thus critical for image-level detection, while localization relies more evenly on both attention and MLP features.

\paragraph{JPEG augmentation.} Applying JPEG compression as a data augmentation during training (row~\#4 vs.\ \#6, both $r{=}1$, 5k steps, batch 20, Train) reduces Avg Loc-F1 from 0.572 to 0.564 ($-1\%$) and Avg Det-F1 from 0.595 to 0.585 ($-2\%$). JPEG augmentation introduces compression artifacts that may distract the model from learning forgery-specific traces, and its mild degradation is consistent across both tasks.

\paragraph{Synthetic data and paired training.} Adding synthetic forgeries to real training data (row~\#6 $\rightarrow$ \#8) raises Avg Loc-F1 by $+9\%$ (0.572 $\rightarrow$ 0.625) and Avg Det-F1 by $+9\%$ (0.595 $\rightarrow$ 0.649). Constructing each batch with matched (clean, forged) pairs of the same document (row~\#8 $\rightarrow$ \#9) further lifts Avg Det-F1 from 0.649 to 0.677, while slightly reducing Avg Loc-F1 from 0.625 to 0.619. Relative to the real-only baseline, the paired setting still improves Avg Loc-F1 by $+8\%$ and Avg Det-F1 by $+14\%$ (row~\#6 vs.\ \#9). Synthetic data therefore provides the main gain across both tasks, while explicit clean-forgery pairing yields an additional detection-specific benefit at a small localization cost.

\paragraph{Training image size.} Larger training resolutions consistently improve localization. On Train-1000 at det-thr$=0.99$, pixel F1 rises from 0.673 at 512px to 0.720 at 1024px and 0.737 at 1280px; Syn-1000 shows the same trend (0.636, 0.682, 0.694). Scaling from 512px to 1280px (at det-thr$=0.99$) improves pixel F1 by roughly $9\%$--$10\%$ on both splits, while moving from 1024px to 1280px adds only about $2\%$. Relaxing the detection threshold to 0.95 at 1280px further raises pixel F1 to 0.761 (Train-1000) and 0.725 (Syn-1000), but at a steep cost: 1280px has $6.25\times$ and $1.56\times$ as many pixels as 512px and 1024px, respectively. Larger images recover finer forgery boundaries, but this benefit must be weighed against the higher training cost.

\paragraph{Detection-first inference and threshold calibration.} Table~\ref{tab:det_results} evaluates a two-stage inference strategy where the detection score gates whether the localization mask is used (above threshold) or suppressed (below), eliminating false-positive masks on predicted-authentic images. Raising both the detection and mask thresholds from 0.50 to 0.99 increases pixel precision from 0.314 to 0.889 ($+183\%$) and image-level F1 from 0.853 to 0.992 ($+16\%$), at the cost of pixel recall dropping from 0.760 to 0.541 ($-29\%$); the best pixel F1 is 0.673 at det-thr$=0.99$. Pushing to 0.999 raises precision to 0.980 but collapses recall to 0.324 (pixel F1 0.487).

\subsection{Evolving Harness Results}

\begin{table}[tb]
    \caption{Meta-Harness evolution results on the 50 samples search set. $S_{\text{Exp}}$ denotes BERTScore F1. $S_{\text{Rep}}$ denotes LLM-judge score. Schema denotes report format validity rate.}
    \label{tab:harness_results}
    \centering
    \normalsize
    \setlength{\tabcolsep}{5pt}
    \begin{tabular}{lccc}
        \toprule
        Stage & $S_{\text{Exp}}$ & $S_{\text{Rep}}$ & Schema Val. \\
        \midrule
        Seed harness (iteration~0) & 68.7 & 76.2 & 0.94 \\
        After 5 iterations & 69.8 & 77.5 & 0.95 \\
        After 15 iterations & 71.4 & 78.8 & 0.96 \\
        After 30 iterations (selected) & \textbf{72.4} & \textbf{79.8} & \textbf{0.98} \\
        \bottomrule
    \end{tabular}
\end{table}

Table~\ref{tab:harness_results} shows the Meta-Harness evolution trajectory. The seed template harness already achieves decent performance ($S_{\text{Exp}} = 68.7$, $S_{\text{Rep}} = 76.2$, schema validity 0.94) due to its built-in schema repair. Over 30 iterations, the proposer LLM discovers improvements such as coordinate-span repair for grounding boxes, calibrated risk-score estimation, and evidence-chain prompting, yielding steady gains in both explanation quality and report structure. The final selected harness improves $S_{\text{Exp}}$ by $+3.7$ and $S_{\text{Rep}}$ by $+3.6$ over the seed, with schema validity reaching 0.98. Critically, no human prompt engineering was involved. The entire progression is auto-generated by the Meta-Harness loop.

\section{Conclusion}

We presented \textbf{SEED}, a pipeline for explainable text-centric image forgery analysis that achieved 3rd place in the GenText-Forensics Challenge at ACM MM 2026, combining three modules: contrastive-guided synthetic forgery generation, a LoRA-adapted ViT detector that preserves pre-trained priors while achieving strong cross-domain localization with minimal parameters, and a Meta-Harness that automatically discovers effective MLLM harnesses for structured report generation. Our experiments show that the detector overfits training-set patterns, and that reducing capacity---lower LoRA rank, fewer steps, smaller batches---consistently improves cross-domain generalization. Future work could address MLLM hallucination in forensic reasoning and explore stronger generative models for producing higher-quality forgeries, especially for challenging domains such as RTM.

\begin{acks}
% TODO (REQUIRED before final submission): Replace the placeholder below
% with the actual funding/acknowledgment statement and grant numbers, e.g.:
% This work was supported in part by the Science and Technology Development
% Fund (FDCT) of Macau SAR under Grant No. XXXX and by the University of
% Macau under Grant No. XXXX.
This work was supported in part by Macau Science and Technology Development Fund under 001/2024/SKL, 0119/2024/RIB2 and 0110/2025/R1B2; in part by Research Committee at University of Macau under MYRG-CRG2025-00031-FST and MYRG-GRG2025-00086-FST; in part by the Guangdong Basic and Applied Basic Research Foundation under Grant 2024A1515012536.
\end{acks}

\clearpage
\bibliographystyle{ACM-Reference-Format}
\bibliography{ref}

\end{document}